\def\year{2020}\relax
\documentclass[letterpaper]{article} 
\usepackage{aaai20}  
\usepackage{times}  
\usepackage{helvet} 
\usepackage{courier}  
\usepackage[hyphens]{url}  
\usepackage{graphicx} 
\usepackage{graphicx}  
\usepackage{latexsym}

\usepackage{url}
\usepackage{times}
\usepackage{helvet}
\usepackage{courier}
\usepackage{amsmath, amssymb}
\usepackage{bm}
\usepackage{graphicx}
\usepackage{subfig}
\usepackage{multirow}
\usepackage{algorithm}
\usepackage[noend]{algpseudocode}
\usepackage{enumitem}
\usepackage{fancyvrb}
\usepackage{booktabs}

\title{Latent-Variable Non-Autoregressive Neural Machine Translation with Deterministic Inference Using a Delta Posterior}

\author{Raphael Shu\textsuperscript{\rm 1}, Jason Lee\textsuperscript{\rm 2}, Hideki Nakayama\textsuperscript{\rm 1}, and Kyunghyun Cho\textsuperscript{\rm 2,3,4}\\
\textsuperscript{\rm 1}The University of Tokyo\\
\textsuperscript{\rm 2}New York University\\
\textsuperscript{\rm 3}Facebook AI Research\\
\textsuperscript{\rm 4}CIFAR Azrieli Global Scholar\\
}

\DeclareMathOperator*{\argmax}{argmax}   

\algnewcommand{\Inputs}[1]{%
  \State \textbf{Inputs:}
  \Statex \hspace*{\algorithmicindent}\parbox[t]{.8\linewidth}{\raggedright #1}
}
\algnewcommand{\Initialize}[1]{%
  \State \textbf{Initialize:}
  \Statex \hspace*{\algorithmicindent}\parbox[t]{.8\linewidth}{\raggedright #1}
}
\newcommand*{\Break}{\textbf{break}}

\newcommand{\citet}[1]{\citeauthor{#1} \shortcite{#1}}
\newcommand{\citep}{\cite}

\date{}

\begin{document}
\maketitle
\begin{abstract}
Although neural machine translation models reached high translation quality, the autoregressive nature makes inference difficult to parallelize and leads to high translation latency. Inspired by recent refinement-based approaches, we propose LaNMT, a latent-variable non-autoregressive model with continuous latent variables and deterministic inference procedure. 
In contrast to existing approaches, we use a deterministic  inference algorithm to find the target sequence that maximizes the lowerbound to the log-probability. During inference, the length of translation automatically adapts itself. Our experiments show that the lowerbound can be greatly increased by running the inference algorithm, resulting in significantly improved translation quality. Our proposed model closes the performance gap between non-autoregressive and autoregressive approaches on ASPEC Ja-En dataset with 8.6x faster decoding. On WMT'14 En-De dataset, our model narrows the gap with autoregressive baseline to 2.0 BLEU points with 12.5x speedup. By decoding multiple initial latent variables in parallel and rescore using a teacher model, the proposed model further brings the gap down to 1.0 BLEU point on WMT'14 En-De task with 6.8x speedup.

\end{abstract}

\section{Introduction}

The field of Neural Machine Translation (NMT) has seen significant improvements in recent years~\cite{bahdanau15jointly,wu2016google,Gehring2017ConvolutionalST,Vaswani2017AttentionIA}. Despite impressive improvements in translation accuracy, the autoregressive nature of NMT models have made it difficult to speed up decoding by utilizing parallel model architecture and hardware accelerators. This has sparked interest in \emph{non-autoregressive} NMT models, which predict every target tokens in parallel. In addition to the obvious decoding efficiency, non-autoregressive text generation is appealing as it does not suffer from exposure bias and suboptimal inference.

Inspired by recent work in non-autoregressive NMT using discrete latent variables~\citep{kaiser18fast} and iterative refinement~\citep{lee18deterministic}, we introduce a sequence of continuous latent variables to capture the uncertainty in the target sentence. We motivate such a latent variable model by conjecturing that it is easier to refine lower-dimensional continuous variables\footnote{We use 8-dimensional latent variables in our experiments.} than to refine high-dimensional discrete variables, as done in \citet{lee18deterministic}. Unlike \citet{kaiser18fast}, the posterior and the prior can be jointly trained to maximize the evidence lowerbound of the log-likelihood $\log p(y|x)$.


In this work, we propose a deterministic iterative algorithm to refine the approximate posterior over the latent variables and obtain better target predictions. During inference, we first obtain the initial posterior from a prior distribution $p(z|x)$ and the initial guess of the target sentence from the conditional distribution $p(y|x,z)$. We then alternate between updating the approximate posterior and target tokens with the help of an approximate posterior $q(z|x,y)$. We avoid stochasticity at inference time by introducing a {\it delta posterior} over the latent variables. We empirically find that this iterative algorithm significantly improves the lowerbound and results in better BLEU scores. By refining the latent variables instead of tokens, the length of translation can dynamically adapt throughout this procedure, unlike previous approaches where the target length was fixed throughout the refinement process. In other words, even if the initial length prediction is incorrect, it can be corrected simultaneously with the target tokens.

Our models\footnote{Our code can be found in https://github.com/zomux/lanmt .} outperform the autoregressive baseline on ASPEC Ja-En dataset with 8.6x decoding speedup and bring the performance gap down to 2.0 BLEU points on WMT'14 En-De with 12.5x decoding speedup. By decoding multiple latent variables sampled from the prior and rescore using a autoregressive teacher model, the proposed model is able to further narrow the performance gap on WMT'14 En-De task down to 1.0 BLEU point with 6.8x speedup. The contributions of this work can be summarize as follows: 
\begin{enumerate}
    \item We propose a continuous latent-variable non-autoregressive NMT model for faster inference. The model learns identical number of latent vectors as the input tokens. A length transformation mechanism is designed to adapt the number of latent vectors to match the target length.
    \item We demonstrate a principle inference method for this kind of model by introducing a deterministic inference algorithm. We show the algorithm converges rapidly in practice and is capable of improving the translation quality by around 2.0 BLEU points.
\end{enumerate}

\section{Background}

\subsection{Autoregressive NMT}

In order to model the joint probability of the target tokens $y_1, \cdots, y_{|y|}$ given the source sentence $x$, most NMT models use an autoregressive factorization of the joint probability which has the following form:
\begin{align}
	\log p(y|x) = \sum_{i=1}^{|y|} \log p(y_i|y_{<i}, x), \label{eq:armodel}
\end{align}
where $y_{<i}$ denotes the target tokens preceding $y_i$. Here, the probability of emitting each token $p(y_i|y_{<i}, x)$ is parameterized with a neural network.

To obtain a translation from this model, one could predict target tokens sequentially by greedily taking {\it argmax} of the token prediction probabilities. The decoding progress ends when a ``\texttt{</s>}'' token, which indicates the end of a sequence, is selected. In practice, however, this greedy approach yields suboptimal sentences, and \emph{beam search} is often used to decode better translations by maintaining multiple hypotheses. However, decoding with a large beam size significantly decreases translation speed.

\subsection{Non-Autoregressive NMT}

Although autoregressive models achieve high translation quality through recent advances in NMT, the main drawback is that autoregressive modeling forbids the decoding algorithm to select tokens in multiple positions simultaneously. This results in inefficient use of computational resource and increased translation latency.

In contrast, non-autoregressive NMT models predict target tokens without depending on preceding tokens, depicted by the following objective:
\begin{align}
    \log p(y|x) = \sum_{i=i}^{|y|} \log p(y_i|x). \label{eq:nonauto-object}
\end{align}
As the prediction of each target token $y_i$ now depends only on the source sentence $x$ and its location $i$ in the sequence, the translation process can be easily parallelized. We obtain a target sequence by applying {\it argmax} to all token probabilities.

The main challenge of non-autoregressive NMT is on capturing dependencies among target tokens. As the probability of each target token does not depend on the surrounding tokens, applying {\it argmax} at each position $i$ may easily result in an inconsistent sequence, that includes duplicated or missing words. It is thus important for non-autoregressive models to apply techniques to ensure the consistency of generated words.

\section{Latent-Variable Non-Autoregressive NMT}

In this work, we propose LaNMT, a latent-variable non-autoregressive NMT model by introducing a sequence of continuous latent variables to model the uncertainty about the target sentence. These latent variables $z$ are constrained to have the same length as the source sequence, that is, $|z| = |x|$. Instead of directly maximizing the objective function in Eq.~\eqref{eq:nonauto-object}, we maximize a lowerbound to the marginal log-probability $\log p(y|x)=\log \int p(y|z,x)p(z|x) dz$:
\begin{align}
	\mathcal{L}(\omega, \phi, \theta) = \: &
		\mathbb{E}_{z \sim q_\phi} \big[ \log p_\theta(y|x,z) \big] \nonumber \\
		& - \mathrm{KL}\big[ q_\phi(z|x,y) || p_\omega (z|x) \big],\label{eq:lowerbound} 
\end{align}
where $p_\omega (z|x)$ is the prior, $q_\phi(z|x,y)$ is an approximate posterior and $p_\theta(y|x,z)$ is the decoder. The objective function in Eq.~\eqref{eq:lowerbound} is referred to as the evidence lowerbound (ELBO). As shown in the equation, the lowerbound is parameterized by three sets of parameters: $\omega$, $\phi$ and $\theta$. 

Both the prior $p_\omega$ and the approximate posterior $q_\phi$ are modeled as spherical Gaussian distributions. The model can be trained end-to-end with the reparameterization trick \cite{Kingma2014AutoEncodingVB}.

\begin{figure}[t]
  \centering
  \includegraphics[width=0.90\columnwidth]{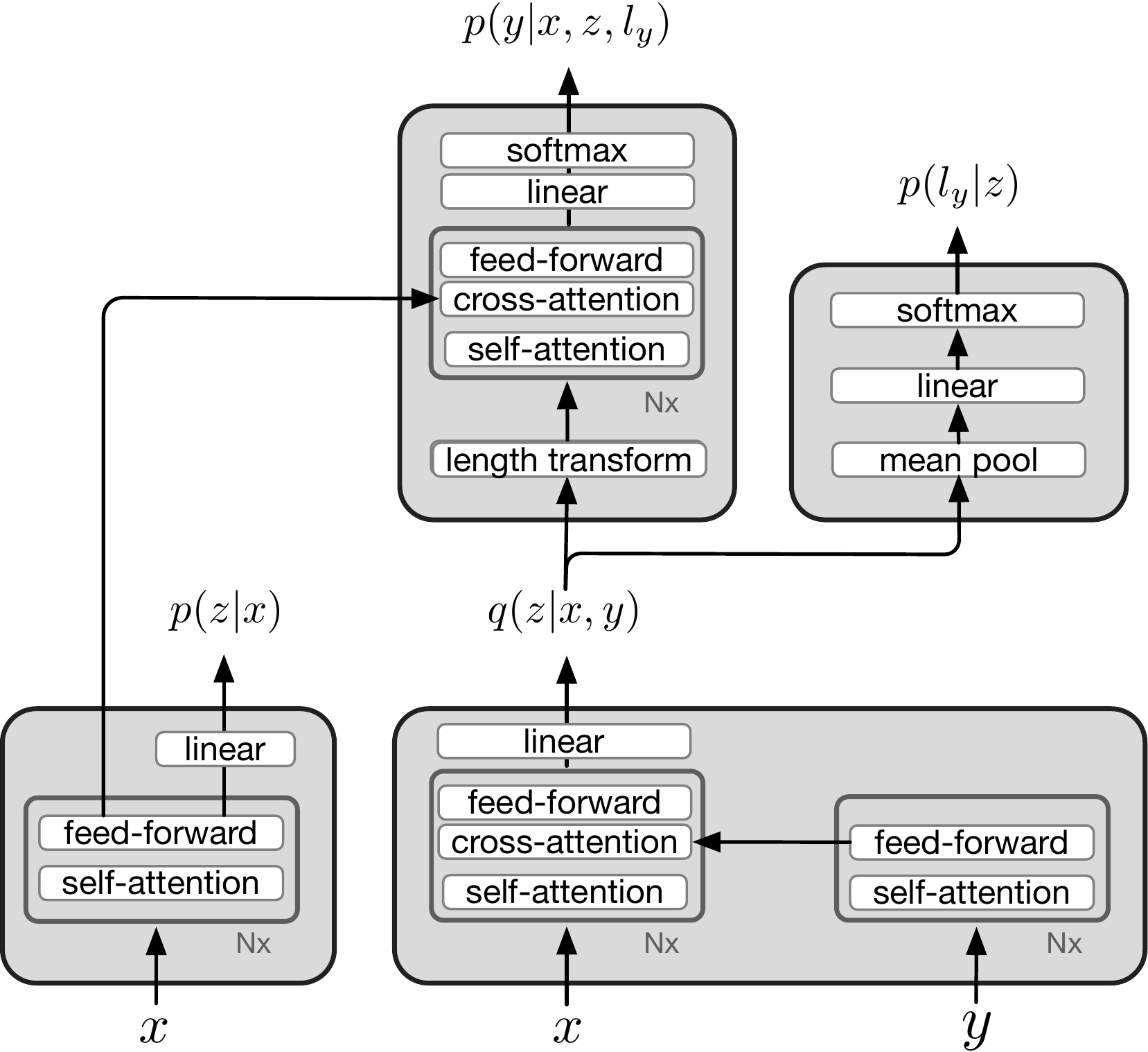}
  \caption{Architecture of the proposed non-autogressive model. The model is composed of four components: prior $p(z|x)$, approximate posterior $q(z|x,y)$, length predictor $p(l_y|z)$ and decoder $p(y|x,z)$. These components are trained end-to-end to maximize the evidence lowerbound.}
  \label{fig:arch}
\end{figure}

\subsection{A Modified Objective Function with Length Prediction}

During training, we want the model to maximize the lowerbound  in Eq.~\eqref{eq:lowerbound}. However, to generate a translation, the target length $l_y$ has to be predicted first. We let the latent variables model the target length by parameterizing the decoder as:
\begin{align}
	p_\theta(y|x,z) &= \sum_l p_\theta(y, l|x,z) \nonumber \\
	&= p_\theta(y, l_y|x, z) \nonumber \\
	&= p_\theta(y|x, z, l_y) p_\theta(l_y|z) \label{eq:likelihood}.
\end{align}

Here $l_y$ denotes the length of $y$. The second step is valid as the probability $p_\theta(y, l \neq l_y|x,z)$ is always zero. Plugging in Eq.~\eqref{eq:likelihood}, with the independent assumption on both latent variables and target tokens, the objective has the following form:
\begin{align}
	\mathbb{E}_{z \sim q_\phi}& \big[ \sum_{i=1}^{|y|}\log p_\theta(y_i|x,z, l_y) + \log p_\theta(l_y|z) \big] \nonumber \\
		& - \sum_{k=1}^{|x|}\mathrm{KL}\big[ q_\phi(z_k|x,y) || p_\omega (z_k|x) \big]. \label{eq:training_loss}
\end{align}

\subsection{Model Architecture}

As evident from in Eq.~\eqref{eq:training_loss}, there are four parameterized components in our model: the prior $p_\omega (z|x)$, approximate posterior $q_\phi(z|x,y)$, decoder $p_\theta(y|x,z, l_y)$ and length predictor $p_\theta(l_y|z)$. The architecture of the proposed non-autoregressive model is depicted in Fig.~\ref{fig:arch}, which reuses modules in Transformer~\citep{Vaswani2017AttentionIA} to compute the aforementioned distributions.

\paragraph{Main Components} To compute the prior $p_\omega(z|x)$, we use a multi-layer self-attention encoder which has the same structure as the Transformer encoder. In each layer, a feed-forward computation is applied after the self-attention. To obtain the probability, we apply a linear transformation to reduce the dimensionality and compute the mean and variance vectors.

For the approximate posterior $q_\phi(z|x, y)$, as it is a function of the source $x$ and the target $y$, we first encode $y$ with a self-attention encoder. Then, the resulting vectors are fed into an attention-based decoder initialized by $x$ embeddings. Its architecture is similar to the Transformer decoder except that no causal mask is used. Similar to the prior, we apply a linear layer to obtain the mean and variance vectors.

To backpropagate the loss signal of the decoder to $q_\phi$, we apply the reparameterization trick to sample $z$ from $q_\phi$ with $g(\epsilon, q) = \mu_q + \sigma_q * \epsilon$. Here, $\epsilon \sim \mathcal{N}(0, 1)$ is Gaussian noise.

The decoder computes the probability $p_{\theta}(y|x,z,l_y)$ of outputting target tokens $y$ given the latent variables sampled from $q_\phi(z|x,y)$. The computational graph of the decoder is also similar to the Transformer decoder without using causal mask. To combine the information from the source tokens, we reuse the encoder vector representation created when computing the prior. 

\begin{figure}[t]
  \centering
  \includegraphics[width=0.56\columnwidth]{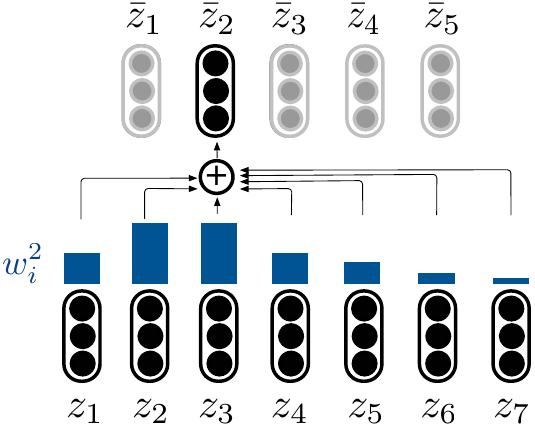}
  \caption{Illustration of the length transformation mechanism.}
  \label{fig:lentransform}
\end{figure}

\paragraph{Length Prediction and Transformation} Given a latent variable $z$ sampled from the approximate posterior $q_\phi$, we train a length prediction model $p_\theta(l_y|z)$. We train the model to predict the length difference between $|y|$ and $|x|$. In our implementation, $p_\theta(l_y|z)$ is modeled as a categorical distribution that covers the length difference in the range $[-50, 50]$. The prediction is produced by applying softmax after a linear transformation.

As the latent variable $z \sim q_\phi(z|x,y)$ has the length $|x|$, we need to transform the latent variables into $l_y$ vectors for the decoder to predict target tokens. We use a monotonic location-based attention for this purpose, which is illustrated in Fig.~\ref{fig:lentransform}. Let the resulting vectors of length transformation be $\bar z_1, ..., \bar z_{l_y}$. we produce each vector with
\begin{align}
	\bar z_j =& \sum_{k=1}^{|x|} w^j_k z_k , \\
	w^j_k =& \frac{\exp ({a^j_k})}{\sum_{k^\prime=1}^{|x|} \exp (a^j_{k^\prime})} ,\\
	a^j_k =& - \frac{1}{2 \sigma ^2} (k - \frac{|x|}{l_y} j)^2,
\end{align}
where each transformed vector is a weighted sum of the latent variables. The weight is computed with a softmax over distance-based logits. We give higher weights to the latent variables close to the location~$\frac{|x|}{l_y}j$. The scale $\sigma$ is the only trainable parameter in this monotonic attention mechanism.

\subsection{Training}

If we train a model with the objective function in Eq.~\eqref{eq:training_loss}, the KL divergence often drops to zero from the beginning. This yields a degenerate model that does not use the latent variables at all. This is a well-known issue in variational inference called posterior collapse~\citep{bowman2015generating,Dieng2018AvoidingLV,Razavi2019PreventingPC}. We use two techniques to address this issue. Similarly to \citet{Kingma2016ImprovingVI}, we give a budget to the KL term as
\begin{align}
	\sum_{k=1}^{|x|}\max(b, \mathrm{KL}\big[ q_\phi(z_k|x,y) || p_\omega (z_k|x) \big]),
\end{align}
where $b$ is the budget of KL divergence for each latent variable. Once the KL value drops below $b$, it will not be minimized anymore, thereby letting the optimizer focus on the reconstruction term in the original objective function. As $b$ is a critical hyperparameter, it is time-consuming to search for a good budget value. Here, we use the following annealing schedule to gradually lower the budget:
\begin{align}
	b = 
    \begin{cases}
    1,&\text{ if } s < M/2 \\
    \frac{(M - s)}{M/2} ,&\text{ otherwise}
    \end{cases}
\end{align}
$s$ is the current step in training, and $M$ is the maximum step. In the first half of the training, the budget $b$ remains $1$. In the second half of the training, we anneal $b$ until it reaches $0$.

Similarly to previous work on non-autoregressive NMT, we apply sequence-level knowledge distillation~\citep{kim16sequence} where we use the output from an autoregressive model as target for our non-autoregressive model.

\section{Inference with a Delta Posterior}

Once the training has converged, we use an inference algorithm to find a translation $y$ that maximizes the lowerbound in Eq.~\eqref{eq:lowerbound}:
\begin{align*}
    \argmax_y \mathbb{E}_{z \sim q_\phi} &\big[ \log p_\theta(y|x,z) \big] 
    \\
    &- \mathrm{KL}\big[ q_\phi(z|x,y) || p_\omega (z|x)
    \big]
\end{align*}
It is intractable to solve this problem exactly due to the intractability of computing the first expectation. We avoid this issue in the training time by reparametrization-based Monte Carlo approximation. However, it is desirable to avoid stochasticity at inference time where our goal is to present a single most likely target sentence given a source sentence. 

We tackle this problem by introducing a proxy distribution $r(z)$ defined as
\begin{align*}
    r(z) = 
    \begin{cases}
    1,&\text{ if }z = \mu \\
    0,&\text{ otherwise}
    \end{cases}
\end{align*}
This is a Dirac measure, and we call it a {\it delta posterior} in our work. We set this delta posterior to minimize the KL divergence against the approximate posterior $q_{\phi}$, which is equivalent to
\begin{align}
\label{eq:update_r}
&\nabla_{\mu} \log q_{\phi}(\mu|x,y) = 0 
\Leftrightarrow \mu = \mathbb{E}_{q_{\phi}}\left[ z \right].
\end{align}

We then use this proxy instead of the original approximate posterior to obtain a {\it deterministic lowerbound}:
\begin{align*}
    \hat{\mathcal{L}}(\omega, \theta, \mu)
    = 
    \log p_{\theta} (y|x, z=\mu)
    - 
    \log p_{\omega} (\mu |x).
\end{align*}
As the second term is constant with respect to $y$, maximizing this lowerbound with respect to $y$ reduces to
\begin{align}
    \label{eq:update_y}
    \argmax_y \log p_{\theta}(y | x, z=\mu),
\end{align}
which can be approximately solved by beam search when $p_\theta$ is an autoregressive sequence model. If $p_\theta$ factorizes over the sequence $y$, as in our non-autoregressive model, we can solve it exactly by
\begin{align*}
    \hat y_i = \argmax_{y_i} \log p_{\theta}(y_i | x, z=\mu).
\end{align*}
With every estimation of $y$, the approximate posterior $q$ changes. We thus alternate between fitting the delta posterior in Eq.~\eqref{eq:update_r} and finding the most likely sequence $y$ in Eq.~\eqref{eq:update_y}.

We initialize the delta posterior $r$ using the prior distribution:
\begin{align*}
    \mu = \mathbb{E}_{p_\omega(z|x)}\left[ z \right].
\end{align*}
With this initialization, the proposed inference algorithm is fully deterministic. The complete inference algorithm for obtaining the final translation is shown in Algorithm~\ref{alg}.




\begin{algorithm}[t]
\caption{Deterministic Iterative Inference}
\label{alg}
\begin{algorithmic}
\Inputs{
    $x :$ source sentence \\
    $T :$ maximum step \\
}
\State $\mu_0 = \mathbb{E}_{p_\omega(z|x)}\left[ z \right]$
\State $y_0 = \argmax_{y} \log p_\theta(y|x, z=\mu_0)$
\For{$t \gets 1 \textrm{ to } T$}
    \State $\mu_t = \mathbb{E}_{q_\phi(z|x, y_{t-1})}\left[z\right]$
    \State $y_t = \argmax_{y} \log p_\theta(y|x, z=\mu_t)$ 
    \If{$y_t = y_{t-1}$}
    \State \Break
    \EndIf
\EndFor
\State \textbf{output} $y_t$
\end{algorithmic}
\end{algorithm}

\section{Related Work}

This work is inspired by a recent line of work in non-autoregressive NMT. \citet{Gu2018NonAutoregressiveNM} first proposed a non-autoregressive framework by modeling word alignment as a latent variable, which has since then been improved by \citet{Wang2019NonAutoregressiveMT}. \citet{lee18deterministic} proposed a deterministic iterative refinement algorithm where a decoder is trained to refine the hypotheses. Our approach is most related to \citet{kaiser18fast,Roy2018TheoryAE}. In both works, a discrete autoencoder is first trained on the target sentence, then an autoregressive prior is trained to predict the discrete latent variables given the source sentence. Our work is different from them in three ways: (1) we use continuous latent variables and train the approximate posterior $q(z|x,y)$ and the prior $p(z|x)$ jointly; (2) we use a non-autoregressive prior; and (3) the refinement is performed in the latent space, as opposed to discrete output space (as done in most previous works using refinement for non-autoregressive machine translation).

Concurrently to our work, \citet{Ghazvininejad2019ConstantTimeMT} proposed to translate with a masked-prediction language model by iterative replacing tokens with low confidence. \citet{gu2019insertion,stern2019insertion,welleck2019non} proposed insertion-based NMT models that insert words to the translations with a specific strategy. Unlike these works, our approach performs refinements in the low-dimensional latent space, rather than in the high-dimensional discrete space.

Similarly to our latent-variable model, \citet{Zhang2016VariationalNM} proposed a variational NMT, and \citet{Shah2018GenerativeNM} and \citet{Eikema2018AutoEncodingVN} models the joint distribution of source and target. Both of them use autoregressive models. \citet{Shah2018GenerativeNM} designed an EM-like algorithm similar to Markov sampling \cite{Arulkumaran2017ImprovingSF}. In contrast, we propose a deterministic algorithm to remove any non-determinism during inference.


\begin{table*}[t]
  \begin{center}
    \begin{tabular}{r|rrr|rrr} 
      \small{} & \multicolumn{3}{c|}{ASPEC Ja-En} & \multicolumn{3}{c}{WMT'14 En-De} \\
      {} &  {BLEU(\%)} &  {speedup} &  {wall-clock (std)} & {BLEU(\%)} &  {speedup} &  {wall-clock (std)} \\
      \toprule
      {Base Transformer, beam size=3} & 27.1 & 1x & 415ms (159) & 26.1 & 1x & 602ms (274) \\
      {Base Transformer, beam size=1} & 24.6 & 1.1x & 375ms (150) & 25.6 &  1.3x & 461ms (219) \\
      \midrule
      {Latent-Variable NAR Model} & 13.3 & 17.0x & 24ms (2) & 11.8 & 22.2x & 27ms (1)  \\
      {+ knowledge distillation} & 25.2 & 17.0x & 24ms (2) & 22.2 & 22.2x & 27ms (1)\\
      {+ deterministic inference} & 27.5 & 8.6x & 48ms (2) & 24.1 & 12.5x & 48ms (8)\\
      {+ latent search} & 28.3 & 4.8x & 86ms (2) & 25.1 & 6.8x & 88ms (8) \\
    \end{tabular}
    \caption{Comparison of the proposed non-autoregressive (NAR) models with the autoregressive baselines. Our implementation of the Base Transformer is 1.0 BLEU point lower than the original paper~\cite{Vaswani2017AttentionIA} on WMT'14 dataset.}
    \label{table:main}
  \end{center}
\end{table*}


\section{Experimental Settings}

\paragraph{Data and preprocessing}

We evaluate our model on two machine translation datasets: ASPEC Ja-En~\citep{NAKAZAWA16.621} and WMT'14 En-De~\citep{bojar-EtAl:2014:W14-33}. The ASPEC dataset contains 3M sentence pairs, and the WMT'14 dataset contains 4.5M senence pairs.

To preprocess the ASPEC dataset, we use Moses toolkit~\citep{Koehn2007MosesOS} to tokenize the English sentences, and Kytea~\citep{kytea} for Japanese sentences. We further apply byte-pair encoding~\citep{rico2106bpe} to segment the training sentences into subwords. The resulting vocabulary has 40K unique tokens on each side of the language pair.
To preprocess the WMT'14 dataset, we apply sentencepiece \cite{Kudo2018SentencePieceAS} to both languages to segment the corpus into subwords and build a joint vocabulary. The final vocabulary size is 32K for each language.

\paragraph{Learning} 

To train the proposed non-autoregressive models, we adapt the same learning rate annealing schedule as the Base Transformer. Model hyperparameters are selected based on the validation ELBO value.

The only new hyperparameter in the proposed model is the dimension of each latent variable. If each latent is a high-dimension vector, although it has a higher capacity, the KL divergence in Eq.~\eqref{eq:lowerbound} becomes difficult to minimize. In practice, we found that latent dimensionality values between 4 and 32 result in similar performance. However, when the dimensionality is significantly higher or lower, we observed a performance drop. In all experiments, we set the latent dimensionionality to 8. We use a hidden size of 512 and feedforward filter size of 2048 for all models in our experiments. We use 6 transformer layers for the prior and the decoder, and 3 transformer layers for the approximate posterior. 

\paragraph{Evaluation}

We evaluate the {\it tokenized BLEU} for ASPEC Ja-En datset. For WMT'14 En-De datset, we use SacreBLEU \cite{post-2018-call} to evaluate the translation results. We follow \citet{lee18deterministic} to remove repetitions from the translation results before evaluating BLEU scores.




\paragraph{Latent Search}

To further exploit the parallelizability of GPUs, we sample multiple initial latent variables from the prior $p_{\omega}(z|x)$. Then we perform the deterministic inference on each latent variable to obtain a list of candidate translations. However, we can not afford to evaluate each candidate using Eq.~\eqref{eq:training_loss}, which requires importance sampling on $q_\phi$. Instead, we use the autoregressive baseline model to score all the candidates, and pick the candidate with the highest log probability. Following \citet{Parmar2018ImageT}, we reduce the temperature by a factor of $0.5$ when sampling latent variables, resulting in better translation quality. To avoid stochasticity, we fix the random seed during sampling. 

\section{Result and Analysis}

\subsection{Quantitative Analysis}

Our quantitative results on both datasets are presented in Table \ref{table:main}. The baseline model in our experiments is a base Transformer. Our implementation of the autoregressive baseline is 1.0 BLEU points lower than the original paper \cite{Vaswani2017AttentionIA} on WMT'14 En-De datase. We measure the latency of decoding each sentence on a single NVIDIA V100 GPU for all models, which is averaged over all test samples.

As shown in Table~\ref{table:main}, without knowledge distillation, we observe a significant gap in translation quality compared to the autoregressive baseline. This observation is in line with previous works on non-autoregressive NMT \cite{Gu2018NonAutoregressiveNM,lee18deterministic,Wang2019NonAutoregressiveMT}. The gap is significantly reduced by using knowledge distillation, as translation targets provided by the autoregressive model are easier to predict.

With the proposed deterministic inference algorithm, we significantly improve translation quality by 2.3 BLEU points on ASPEC Ja-En dataset and 1.9 BLEU points on WMT'14 En-De dataset. Here, we only run the algorithm for one step. We observe gain on ELBO by running more iterative steps, which is however not reflected by the BLEU scores. As a result, we outperform the autoregressive baseline on ASPEC dataset with a speedup of 8.6x. For WMT'14 dataset, although the proposed model reaches a speedup of 12.5x, the gap with the autoregressive baseline still remains, at 2.0 BLEU points. We conjecture that WMT'14 En-De is more difficult for our non-autoregressive model as it contains a high degree of noise~\citep{ott18analyzing}.

By searching over multiple initial latent variables and rescoring with the teacher Transformer model, we observe an increase in performance by $0.7 \sim 1.0$ BLEU score at the cost of lower translation speed. In our experiments, we sample 50 candidate latent variables and decode them in parallel. The slowdown is mainly caused by rescoring. With the help of rescoring, our final model further narrows the performance gap with the autoregressive baseline to 1.0 BLEU with 6.8x speedup on WMT'14 En-De task.
 
\subsection{Non-autoregressive NMT Models}

In Table~\ref{table:nar-compare}, we list the results on WMT'14 En-De by existing non-autoregressive NMT approaches. All the models use Transformer as their autoregressive baselines. In comparison, our proposed model suffers a drop of 1.0 BLEU points over the baseline, which is a relatively small gap among the existing models.
Thanks to the rapid convergence of the proposed deterministic inference algorithm, our model achieves a higher speed-up compared to other refinement-based models and provides a better speed-accuracy tradeoff.

Concurrently to our work, the mask-prediction language model \cite{Ghazvininejad2019ConstantTimeMT} was found to reduce the performance gap down to 0.9 BLEU on WMT'14 En-De while still maintaining a reasonable speed-up. The main difference is that we update a delta posterior over latent variables instead of target tokens.
Both \citet{Ghazvininejad2019ConstantTimeMT} and \citet{Wang2019NonAutoregressiveMT} with autoregressive rescoring decode multiple candidates in batch and choose one final translation from them. FlowSeq \cite{Ma2019FlowSeqNC} is an recent interesting work on flow-based prior. With noisy parallel decoding, FlowSeq can be fairly compared to the latent search setting of our model. In Table \ref{table:nar-compare}, we can see that our model is equivalently or more effective without a flow-based prior. It is intriguing to see a combination with the flow approach.

\begin{table}[t]
  \begin{center}
    \begin{tabular}{r|lr}
      {} &  \small{BLEU(\%)} &  \small{speed-up} \\
      \toprule
      \small{Transformer \cite{Vaswani2017AttentionIA}} & 27.1 & -\\
      \midrule
      \small{Baseline \cite{Gu2018NonAutoregressiveNM}} & 23.4 & 1x\\
      \small{NAT (+FT +NPD S=100)} & 19.1 \small{(-4.3)} & 2.3x\\
      \midrule
      \small{Baseline \cite{lee18deterministic}} & 24.5 & 1x\\
      \small{Adaptive NAR Model} & 21.5 \small{(-3.0)}& 1.9x\\
      \midrule
      \small{Baseline \cite{kaiser18fast}} & 23.5 & 1x\\
      \small{LT, Improved Semhash} & 19.8 \small{(-3.7)}& 3.8x\\
      \midrule
      \small{Baseline \cite{Wang2019NonAutoregressiveMT}} & 27.3 & 1x\\
      \small{NAT-REG, no rescoring} & 20.6 \small{(-6.7)}& \small{27.6x$^\star$}\\
      \small{NAT-REG, autoregressive rescoring} & 24.6 \small{(-2.7)}& \small{15.1x$^\star$}\\
      \midrule
      \small{Baseline \cite{Ghazvininejad2019ConstantTimeMT}} & 27.8 & 1x\\
      \small{CMLM with 4 iterations} & 26.0 \small{(-1.8)} & -\\
      \small{CMLM with 10 iterations} & 26.9 \small{(-0.9)}& \small{2$\sim$3x}\\
      \midrule
      \small{Baseline \cite{Ma2019FlowSeqNC}} & 27.1 & - \\
      \small{FlowSeq-large (NPD n = 30)} & 25.3 (-1.8) & - \\
      \midrule
      \small{Baseline (Ours)} & 26.1 & 1x\\
      \small{NAR with deterministic Inference} & 24.1 \small{(-2.0)} & \small{12.5x}\\
      \small{+ latent search} & 25.1 \small{(-1.0)} & \small{6.8x}\\
    \end{tabular}
    \caption{A comparison of non-autoregressive NMT models on WMT'14 En-De dataset in BLEU(\%) and decoding speed-up. $\star$ measured on IWSLT'14 DE-EN dataset.}
    \label{table:nar-compare}
  \end{center}
\end{table}


\begin{figure}[t]
  \centering
  \includegraphics[width=0.92\columnwidth]{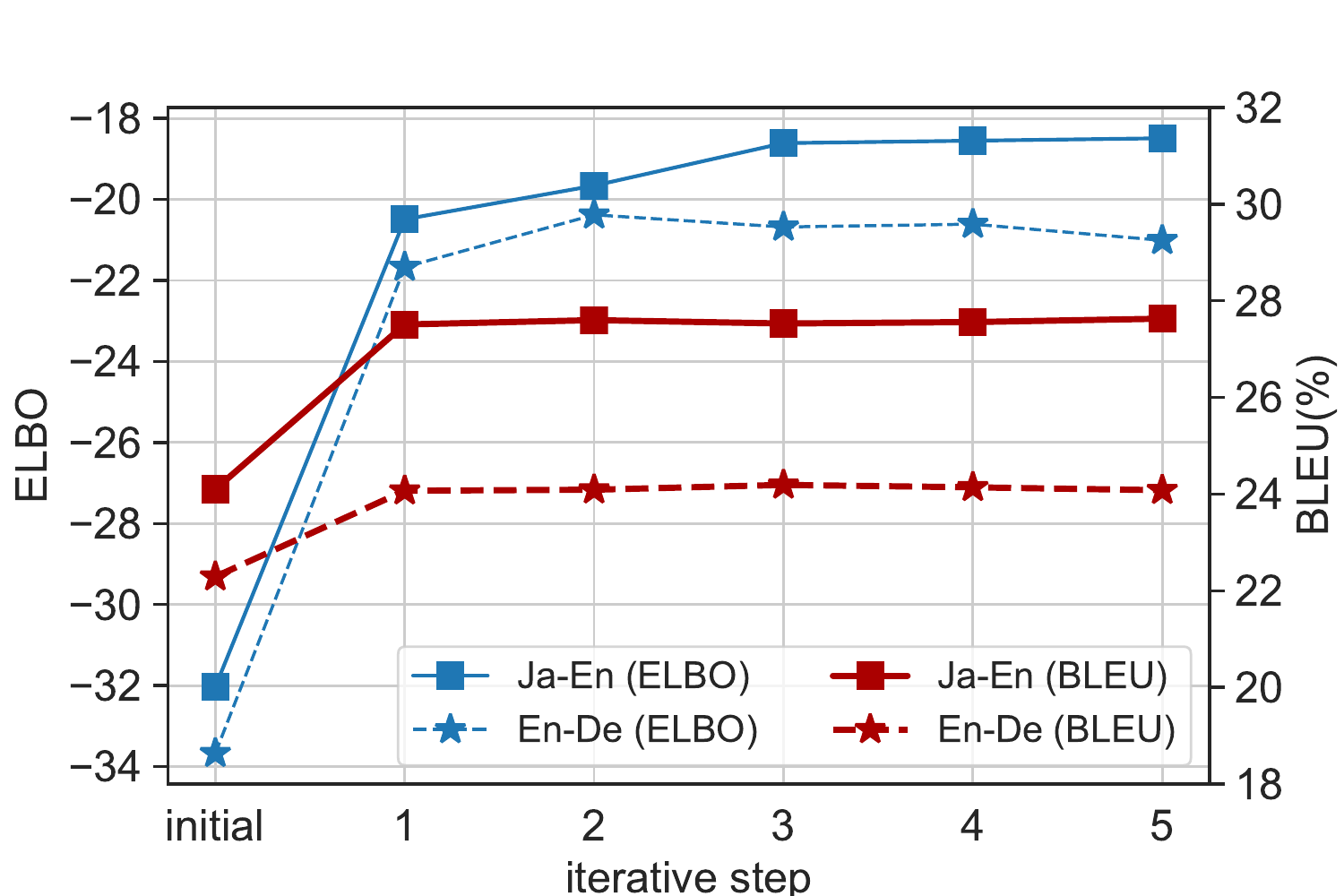}
  \caption{ELBO and BLEU scores measured with the target predictions obtained at each inference step for ASPEC Ja-En and WMT'14 En-De datasets. }
  \label{fig:elbo}
\end{figure}

\begin{figure}[t]
  \centering
  \includegraphics[width=0.96\columnwidth]{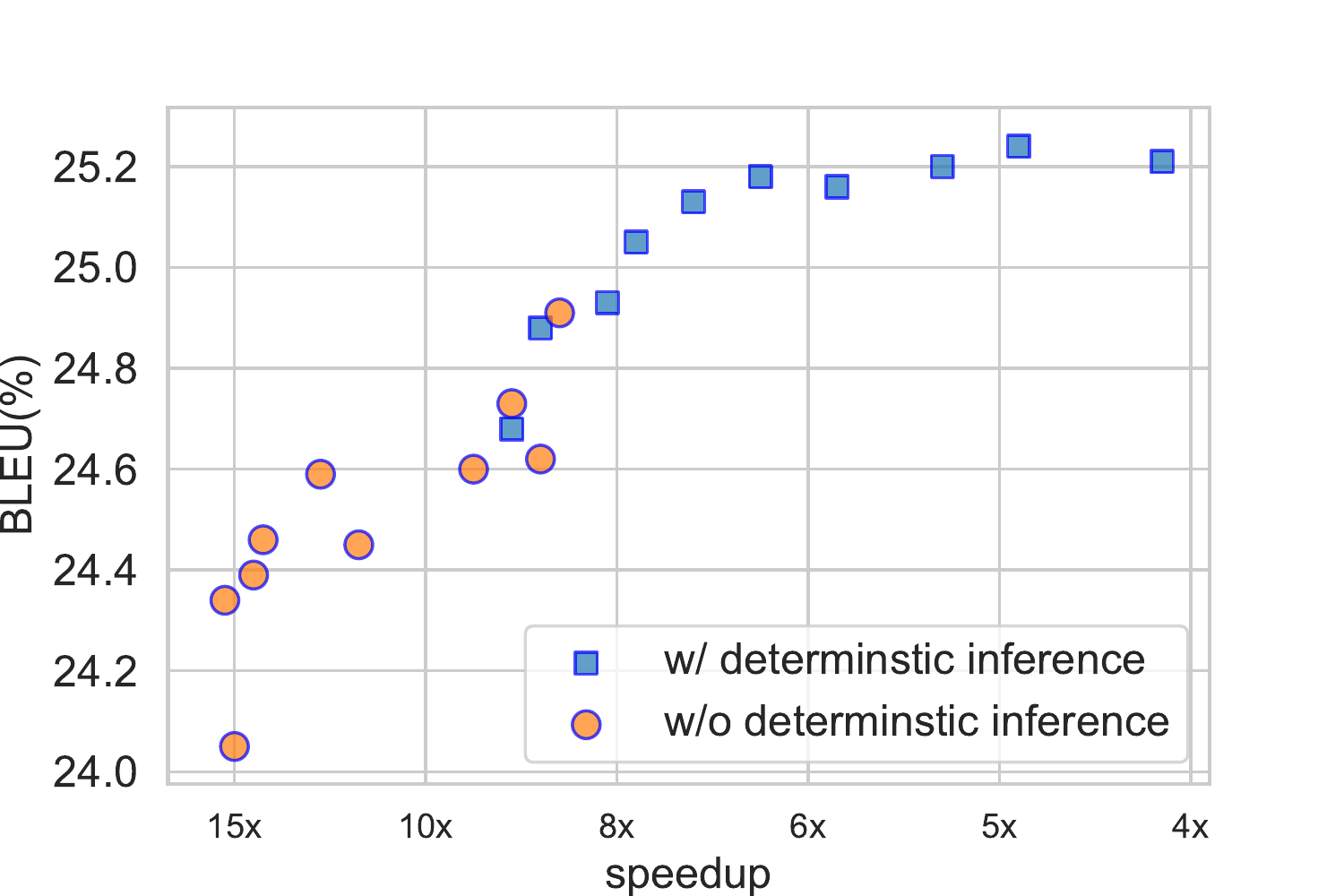}
    \caption{Trade-off between BLEU scores and speedup on WMT'14 En-De task by varying the number of candidates computed in parallel from 10 to 100.}
  \label{fig:tradeoff}
\end{figure}

\subsection{Analysis of Deterministic Inference}
\label{section:convergence}

\paragraph{Convergences of ELBO and BLEU}

In this section, we empirically show that the proposed deterministic iterative inference improves the ELBO in Eq.~\eqref{eq:lowerbound}.
As the ELBO is a function of $x$ and $y$, we measure the ELBO value with the new target prediction after each iteration during inference. For each instance, we sample 20 latent variables to compute the expectation in Eq.~\eqref{eq:lowerbound}. The ELBO value is further averaged over data samples. 

In Fig.~\ref{fig:elbo}, we show the ELBO value and the resulting BLEU scores for both datasets. In the initial step, the delta posterior is initialized with the prior distribution $p_\omega(z|x)$. We see that the ELBO value increases rapidly with each refinement step, which means a higher lowerbound to $\log p(y|x)$. The improvement is highly correlated with increasing BLEU scores. For around 80\% of the data samples, the algorithm converges within three steps. We observe the BLEU scores peaked after only one refinement step.

\begin{table*}[t]
  \begin{center}
    \begin{tabular}{r|l} 
      \multicolumn{2}{l}{Example 1: Sequence modified without changing length} \\
      \toprule

      {Source} & \small{\texttt{hyouki gensuiryou hyoujun no kakuritsu wo kokoromita. (Japanese)}} \\
      {Reference} & \small{\texttt{the establishment of an optical fiber attenuation standard was attempted .}} \\
      {Initial Guess} & \small{\texttt{an attempt was made establish establish damping attenuation standard ...}} \\
      {After Inference} & \small{\texttt{an attempt was \underline{to establish the} damping attenuation standard ...}} \\

      \multicolumn{2}{c}{} \\ 
      \multicolumn{2}{l}{Example 2: One word removed from the sequence} \\
      \toprule
      {Source} & \small{\texttt{...``sen bouchou keisu no toriatsukai'' nitsuite nobeta. (Japanese)}} \\
      {Reference} & \small{\texttt{... handling of linear expansion coefficient .}} \\
      {Initial Guess} & \small{\texttt{... `` handling of of linear expansion coefficient '' are described .}} \\
      {After Inference} & \small{\texttt{... `` handling \underline{of linear} expansion coefficient '' are described .}} \\

      \multicolumn{2}{c}{} \\ 
      \multicolumn{2}{l}{Example 3: Four words added to the sequence} \\
      \toprule

    {Source} & \small{\texttt{... maikuro manipyureshon heto hatten shite kite ori ...(Japanese)}} \\
      {Reference} & \small{\texttt{... with wide application fields so that it has been developed ...}} \\
      {Initial Guess} & \small{\texttt{... micro micro manipulation and ...}} \\
      {After Inference} & \small{\texttt{... and micro manipulation \underline{, and it has been developed ,} and ...}} \\

      \multicolumn{2}{c}{} \\ 
    \end{tabular}
    \caption{Ja-En sample translation with the proposed iterative inference algorithm. In the first example, the initial guess is refined without a change in length. In the last two examples, the iterative inference algorithm changes the target length along with its content. This is more pronounced in the last example, where a whole clause is inserted during refinement.
    }
    \label{table:examples}
  \end{center}
\end{table*}


\paragraph{Trade-off between Quality and Speed}

In Fig.~\ref{fig:tradeoff}, we show the trade-off between translation quality and the speed gain on WMT'14 En-De task when considering multiple candidates latent variables in parallel. We vary the number of candidates from 10 to 100, and report BLEU scores and relative speed gains in the scatter plot. The results are divided into two groups. The first group of experiments search over multiple latent variables and rescore with the teacher Transformer. The second group applies the proposed deterministic inference before rescoring.

We observe that the proposed deterministic inference consistently improves translation quality in all settings. The BLEU score peaks at 25.2. As GPUs excel at processing massive computations in parallel, we can see that the translation speed only degrades by a small magnitude.

\subsection{Qualitative Analysis}

We present some translation examples to demonstrate the effect of the proposed iterative inference in Table~\ref{table:examples}. In Example 1, the length of the target sequence does not change but only the tokens are replaced over the refinement iterations.
The second and third examples show that the algorithm removes or inserts words during the iterative inference by adaptively changing the target length. Such a significant modification to the predicted sequence mostly happens when translating long sentences. 

For some test examples, however, we still find duplicated words in the final translation after applying the proposed deterministic inference. For them, we notice that the quality of the initial guess of translation is considerably worse than average, which typically contains multiple duplicated words. 
Thus, a high-quality initial guess is crucial for obtaining good translations.

\section{Conclusion}

Our work presents the first approach to use continuous latent variables for non-autoregressive Neural Machine Translation. The key idea is to introduce a sequence of latent variables to capture the uncertainly in the target sentence. The number of latent vectors is always identical to the number of input tokens. A length transformation mechanism is then applied to adapt the latent vectors to match the target length. We train the proposed model by maximizing the lowerbound of the log-probability $\log p(y|x)$.

We then introduce a deterministic inference algorithm that uses a {\it delta posterior} over the latent variables. The algorithm alternates between updating the delta posterior and the target tokens. Our experiments show that the algorithm is able to improve the evidence lowerbound of predicted target sequence rapidly.
In our experiments, the BLEU scores converge in one refinement step. 

Our non-autoregressive NMT model closes the performance gap with autoregressive baseline on ASPEC Ja-En task with a 8.6x speedup. By decoding multiple latent variables sampled from the prior, our model brings down the gap on En-De task down to 1.0 BLEU with a speedup of 6.8x. 



\section*{Acknowledgement}
\begin{small}
We thank eBay and NVIDIA. This work was partly supported by Samsung Advanced Institute of Technology (Next Generation Deep Learning: from pattern recognition to AI), Samsung Electronics (Improving Deep Learning using Latent Structure). The results were achieved by "Research and Development of Deep Learning Technology for Advanced Multilingual Speech Translation", the Commissioned Research of National Institute of Information and Communications Technology (NICT), JAPAN. This work was partially supported by JSPS KAKENHI Grant Number JP19H04166.
\end{small}

\bibliography{mybib}

\begin{thebibliography}{}

\bibitem[\protect\citeauthoryear{Arulkumaran, Creswell, and
  Bharath}{2017}]{Arulkumaran2017ImprovingSF}
Arulkumaran, K.; Creswell, A.; and Bharath, A.~A.
\newblock 2017.
\newblock Improving sampling from generative autoencoders with markov chains.
\newblock {\em CoRR} abs/1610.09296.

\bibitem[\protect\citeauthoryear{Bahdanau, Cho, and
  Bengio}{2015}]{bahdanau15jointly}
Bahdanau, D.; Cho, K.; and Bengio, Y.
\newblock 2015.
\newblock Neural machine translation by jointly learning to align and
  translate.
\newblock In {\em International Conference on Learning Representations}.

\bibitem[\protect\citeauthoryear{Bojar \bgroup et al\mbox.\egroup
  }{2014}]{bojar-EtAl:2014:W14-33}
Bojar, O.; Buck, C.; Federmann, C.; Haddow, B.; Koehn, P.; Leveling, J.; Monz,
  C.; Pecina, P.; Post, M.; Saint-Amand, H.; Soricut, R.; Specia, L.; and
  Tamchyna, A.
\newblock 2014.
\newblock Findings of the 2014 workshop on statistical machine translation.
\newblock In {\em Proceedings of the Ninth Workshop on Statistical Machine
  Translation},  12--58.
\newblock Baltimore, Maryland, USA: Association for Computational Linguistics.

\bibitem[\protect\citeauthoryear{Bowman \bgroup et al\mbox.\egroup
  }{2015}]{bowman2015generating}
Bowman, S.~R.; Vilnis, L.; Vinyals, O.; Dai, A.~M.; Jozefowicz, R.; and Bengio,
  S.
\newblock 2015.
\newblock Generating sentences from a continuous space.
\newblock {\em arXiv preprint arXiv:1511.06349}.

\bibitem[\protect\citeauthoryear{Dieng \bgroup et al\mbox.\egroup
  }{2018}]{Dieng2018AvoidingLV}
Dieng, A.~B.; Kim, Y.; Rush, A.~M.; and Blei, D.~M.
\newblock 2018.
\newblock Avoiding latent variable collapse with generative skip models.
\newblock {\em CoRR} abs/1807.04863.

\bibitem[\protect\citeauthoryear{Eikema and
  Aziz}{2018}]{Eikema2018AutoEncodingVN}
Eikema, B., and Aziz, W.
\newblock 2018.
\newblock Auto-encoding variational neural machine translation.
\newblock In {\em RepL4NLP@ACL}.

\bibitem[\protect\citeauthoryear{Gehring \bgroup et al\mbox.\egroup
  }{2017}]{Gehring2017ConvolutionalST}
Gehring, J.; Auli, M.; Grangier, D.; Yarats, D.; and Dauphin, Y.
\newblock 2017.
\newblock Convolutional sequence to sequence learning.
\newblock {\em CoRR} abs/1705.03122.

\bibitem[\protect\citeauthoryear{Ghazvininejad \bgroup et al\mbox.\egroup
  }{2019}]{Ghazvininejad2019ConstantTimeMT}
Ghazvininejad, M.; Levy, O.; Liu, Y.; and Zettlemoyer, L.~S.
\newblock 2019.
\newblock Constant-time machine translation with conditional masked language
  models.
\newblock {\em CoRR} abs/1904.09324.

\bibitem[\protect\citeauthoryear{Gu \bgroup et al\mbox.\egroup
  }{2018}]{Gu2018NonAutoregressiveNM}
Gu, J.; Bradbury, J.; Xiong, C.; Li, V. O.~K.; and Socher, R.
\newblock 2018.
\newblock Non-autoregressive neural machine translation.
\newblock {\em CoRR} abs/1711.02281.

\bibitem[\protect\citeauthoryear{Gu, Liu, and Cho}{2019}]{gu2019insertion}
Gu, J.; Liu, Q.; and Cho, K.
\newblock 2019.
\newblock Insertion-based decoding with automatically inferred generation
  order.
\newblock {\em arXiv preprint arXiv:1902.01370}.

\bibitem[\protect\citeauthoryear{Kaiser \bgroup et al\mbox.\egroup
  }{2018}]{kaiser18fast}
Kaiser, L.; Roy, A.; Vaswani, A.; Parmar, N.; Bengio, S.; Uszkoreit, J.; and
  Shazeer, N.
\newblock 2018.
\newblock Fast decoding in sequence models using discrete latent variables.
\newblock {\em arXiv preprint arXiv:1803.03382}.

\bibitem[\protect\citeauthoryear{Kim and Rush}{2016}]{kim16sequence}
Kim, Y., and Rush, A.~M.
\newblock 2016.
\newblock Sequence-level knowledge distillation.
\newblock In {\em Proceedings of the 2016 Conference on Empirical Methods in
  Natural Language Processing},  1317--1327.

\bibitem[\protect\citeauthoryear{Kingma and
  Welling}{2014}]{Kingma2014AutoEncodingVB}
Kingma, D.~P., and Welling, M.
\newblock 2014.
\newblock Auto-encoding variational bayes.
\newblock {\em CoRR} abs/1312.6114.

\bibitem[\protect\citeauthoryear{Kingma, Salimans, and
  Welling}{2016}]{Kingma2016ImprovingVI}
Kingma, D.~P.; Salimans, T.; and Welling, M.
\newblock 2016.
\newblock Improving variational inference with inverse autoregressive flow.
\newblock {\em CoRR} abs/1606.04934.

\bibitem[\protect\citeauthoryear{Koehn \bgroup et al\mbox.\egroup
  }{2007}]{Koehn2007MosesOS}
Koehn, P.; Hoang, H.; Birch, A.; Callison-Burch, C.; Federico, M.; Bertoldi,
  N.; Cowan, B.; Shen, W.; Moran, C.; Zens, R.; Dyer, C.; Bojar, O.;
  Constantin, A.; and Herbst, E.
\newblock 2007.
\newblock Moses: Open source toolkit for statistical machine translation.
\newblock In {\em ACL}.

\bibitem[\protect\citeauthoryear{Kudo and
  Richardson}{2018}]{Kudo2018SentencePieceAS}
Kudo, T., and Richardson, J.
\newblock 2018.
\newblock Sentencepiece: A simple and language independent subword tokenizer
  and detokenizer for neural text processing.
\newblock In {\em EMNLP}.

\bibitem[\protect\citeauthoryear{Lee, Mansimov, and
  Cho}{2018}]{lee18deterministic}
Lee, J.; Mansimov, E.; and Cho, K.
\newblock 2018.
\newblock Deterministic non-autoregressive neural sequence modeling by
  iterative refinement.
\newblock In {\em Proceedings of the 2018 Conference on Empirical Methods in
  Natural Language Processing},  1173--1182.

\bibitem[\protect\citeauthoryear{Ma \bgroup et al\mbox.\egroup
  }{2019}]{Ma2019FlowSeqNC}
Ma, X.; Zhou, C.; Li, X.; Neubig, G.; and Hovy, E.~H.
\newblock 2019.
\newblock Flowseq: Non-autoregressive conditional sequence generation with
  generative flow.
\newblock {\em EMNLP}.

\bibitem[\protect\citeauthoryear{Nakazawa \bgroup et al\mbox.\egroup
  }{2016}]{NAKAZAWA16.621}
Nakazawa, T.; Yaguchi, M.; Uchimoto, K.; Utiyama, M.; Sumita, E.; Kurohashi,
  S.; and Isahara, H.
\newblock 2016.
\newblock Aspec: Asian scientific paper excerpt corpus.
\newblock In {\em LREC}.

\bibitem[\protect\citeauthoryear{Neubig, Nakata, and Mori}{2011}]{kytea}
Neubig, G.; Nakata, Y.; and Mori, S.
\newblock 2011.
\newblock Pointwise prediction for robust, adaptable japanese morphological
  analysis.
\newblock In {\em ACL},  529--533.

\bibitem[\protect\citeauthoryear{Ott \bgroup et al\mbox.\egroup
  }{2018}]{ott18analyzing}
Ott, M.; Auli, M.; Grangier, D.; and Ranzato, M.
\newblock 2018.
\newblock Analyzing uncertainty in neural machine translation.
\newblock In {\em Proceedings of the 35th International Conference on Machine
  Learning, {ICML} 2018},  3953--3962.

\bibitem[\protect\citeauthoryear{Parmar \bgroup et al\mbox.\egroup
  }{2018}]{Parmar2018ImageT}
Parmar, N.; Vaswani, A.; Uszkoreit, J.; Kaiser, L.; Shazeer, N.; Ku, A.; and
  Tran, D.
\newblock 2018.
\newblock Image transformer.
\newblock In {\em ICML}.

\bibitem[\protect\citeauthoryear{Post}{2018}]{post-2018-call}
Post, M.
\newblock 2018.
\newblock A call for clarity in reporting {BLEU} scores.
\newblock In {\em Proceedings of the Third Conference on Machine Translation:
  Research Papers},  186--191.
\newblock Belgium, Brussels: Association for Computational Linguistics.

\bibitem[\protect\citeauthoryear{Razavi \bgroup et al\mbox.\egroup
  }{2019}]{Razavi2019PreventingPC}
Razavi, A.; van~den Oord, A.; Poole, B.; and Vinyals, O.
\newblock 2019.
\newblock Preventing posterior collapse with delta-vaes.
\newblock {\em CoRR} abs/1901.03416.

\bibitem[\protect\citeauthoryear{Roy \bgroup et al\mbox.\egroup
  }{2018}]{Roy2018TheoryAE}
Roy, A.; Vaswani, A.; Neelakantan, A.; and Parmar, N.
\newblock 2018.
\newblock Theory and experiments on vector quantized autoencoders.
\newblock {\em CoRR} abs/1805.11063.

\bibitem[\protect\citeauthoryear{Sennrich, Haddow, and
  Birch}{2016}]{rico2106bpe}
Sennrich, R.; Haddow, B.; and Birch, A.
\newblock 2016.
\newblock Neural machine translation of rare words with subword units.
\newblock In {\em ACL},  1715--1725.

\bibitem[\protect\citeauthoryear{Shah and Barber}{2018}]{Shah2018GenerativeNM}
Shah, H., and Barber, D.
\newblock 2018.
\newblock Generative neural machine translation.
\newblock In {\em NeurIPS}.

\bibitem[\protect\citeauthoryear{Stern \bgroup et al\mbox.\egroup
  }{2019}]{stern2019insertion}
Stern, M.; Chan, W.; Kiros, J.; and Uszkoreit, J.
\newblock 2019.
\newblock Insertion transformer: Flexible sequence generation via insertion
  operations.
\newblock {\em arXiv preprint arXiv:1902.03249}.

\bibitem[\protect\citeauthoryear{Vaswani \bgroup et al\mbox.\egroup
  }{2017}]{Vaswani2017AttentionIA}
Vaswani, A.; Shazeer, N.; Parmar, N.; Uszkoreit, J.; Jones, L.; Gomez, A.~N.;
  Kaiser, L.; and Polosukhin, I.
\newblock 2017.
\newblock Attention is all you need.
\newblock In {\em NIPS}.

\bibitem[\protect\citeauthoryear{Wang \bgroup et al\mbox.\egroup
  }{2019}]{Wang2019NonAutoregressiveMT}
Wang, Y.; Tian, F.; He, D.; Qin, T.; Zhai, C.; and Liu, T.-Y.
\newblock 2019.
\newblock Non-autoregressive machine translation with auxiliary regularization.
\newblock {\em CoRR} abs/1902.10245.

\bibitem[\protect\citeauthoryear{Welleck \bgroup et al\mbox.\egroup
  }{2019}]{welleck2019non}
Welleck, S.; Brantley, K.; Daum{\'e}~III, H.; and Cho, K.
\newblock 2019.
\newblock Non-monotonic sequential text generation.
\newblock {\em arXiv preprint arXiv:1902.02192}.

\bibitem[\protect\citeauthoryear{Wu \bgroup et al\mbox.\egroup
  }{2016}]{wu2016google}
Wu, Y.; Schuster, M.; Chen, Z.; Le, Q.~V.; Norouzi, M.; Macherey, W.; Krikun,
  M.; Cao, Y.; Gao, Q.; Macherey, K.; et~al.
\newblock 2016.
\newblock Google's neural machine translation system: Bridging the gap between
  human and machine translation.
\newblock {\em arXiv preprint arXiv:1609.08144}.

\bibitem[\protect\citeauthoryear{Zhang, Xiong, and
  Su}{2016}]{Zhang2016VariationalNM}
Zhang, B.; Xiong, D.; and Su, J.
\newblock 2016.
\newblock Variational neural machine translation.
\newblock In {\em EMNLP}.

\end{thebibliography}
\bibliographystyle{aaai}

\end{document}